\pdfoutput=1

\documentclass[11pt]{article}

\usepackage[preprint]{acl}

\usepackage{times}
\usepackage{latexsym}

\usepackage[T1]{fontenc}

\usepackage[utf8]{inputenc}

\usepackage{microtype}

\usepackage{inconsolata}

\usepackage{graphicx}
\usepackage{amsmath,amsfonts,amssymb,amstext,bm}
\usepackage{bbm}
\usepackage{booktabs,multirow}
\usepackage{url}
\usepackage{subcaption}
\usepackage{tcolorbox}
\usepackage{dashrule}

\usepackage{pgfplots}
\usetikzlibrary{patterns,backgrounds}
\definecolor{r_d}{RGB}{162, 0, 037}
\definecolor{r_l}{RGB}{248, 206, 204}
\definecolor{g_d}{RGB}{0, 87, 0}
\definecolor{g_l}{RGB}{213, 232, 212}
\definecolor{b_d}{RGB}{0, 110, 175}
\definecolor{b_l}{RGB}{218, 232, 252}
\definecolor{o_d}{RGB}{180, 101, 4}
\definecolor{o_l}{RGB}{255, 230, 204}

\title{Contrastive Learning on LLM Back Generation Treebank for \\ Cross-domain Constituency Parsing}

\author{
Peiming Guo\footnotemark[3]\thanks{~~This work was done when Peiming Guo was a research assistant at Westlake University.}, Meishan Zhang\footnotemark[3], Jianling Li\footnotemark[4], Min Zhang\footnotemark[3], Yue Zhang\footnotemark[5]\thanks{~~Corresponding author: Yue Zhang.} \\
\footnotemark[3] Institute of Computing and Intelligence, Harbin Institute of Technology (Shenzhen), China\\
\footnotemark[4] School of New Media and Communication, Tianjin University, China\\
\footnotemark[5] School of Engineering, Westlake University, China\\
\texttt{guopeiming.gpm@gmail.com, mason.zms@gmail.com, jianlingl@tju.edu.cn}\\
\texttt{zhangmin2021@hit.edu.cn, yue.zhang@wias.org.cn}\\
}

\begin{document}
\maketitle
\begin{abstract}
Cross-domain constituency parsing is still an unsolved challenge in computational linguistics since the available multi-domain constituency treebank is limited.
We investigate automatic treebank generation by large language models (LLMs) in this paper.
The performance of LLMs on constituency parsing is poor, therefore we propose a novel treebank generation method, LLM back generation, which is similar to the reverse process of constituency parsing.
LLM back generation takes the incomplete cross-domain constituency tree with only domain keyword leaf nodes as input and fills the missing words to generate the cross-domain constituency treebank.
Besides, we also introduce a span-level contrastive learning pre-training strategy to make full use of the LLM back generation treebank for cross-domain constituency parsing.
We verify the effectiveness of our LLM back generation treebank coupled with contrastive learning pre-training on five target domains of MCTB.
Experimental results show that our approach achieves state-of-the-art performance on average results compared with various baselines.
\end{abstract}

\section{Introduction}
Constituency parsing is a fundamental task in computational linguistics that aims to build a hierarchical syntax tree for the given sentence.
Although chart-based parsers~\citep{stern2017minimal,kitaev-klein-2018-constituency,teng2018two} achieve state-of-the-art results for the in-domain scenario (at least 95\% F1 score for the news domain) based on the supervised learning and large-scale treebanks~\citep{kitaev2019multilingual,zhang2020efficient,tian2020improving,cui2021investigating}, there is a performance gap in out-of-domain settings, since available multi-domain constituency treebank is limited~\citep{yang2022challenges,li-etal-2023-llm,guo2024cross}.
Therefore, stable cross-domain constituency parsing performance is still a challenge for constituency parsing.

Using large language models (LLMs)~\citep{brown2020language,ouyang2022training,openai2023gpt} is a promising solution for dataset annotation in various natural language processing tasks, such as text classification~\citep{tornberg2023chatgpt}, named entity recognition~\citep{zhang-etal-2023-llmaaa}, semantic search~\citep{bansal2023large}, etc.
For cross-domain constituency parsing, \citet{li-etal-2023-llm} first employ ChatGPT to generate unlabeled raw sentences based on grammar rules, and then the chart-based parser annotates a pseudo treebank on them.
Integrating with self-training, this two-stage approach gains state-of-the-art cross-domain constituency parsing performance.

\begin{figure}
    \centering
    \includegraphics[width=0.48\textwidth]{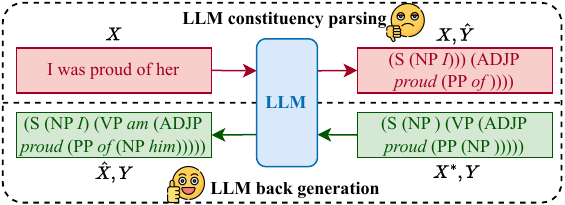}
    \caption{LLM constituency parsing usually predicts the wrong constituency tree structure $\widehat{Y}$ for the input sentence $X$. However, LLM back generation can generate valid constituency tree structure with the appropriate sentence $\widehat{X}$ based on the masked sentence $X^{*}$ and bare constituency tree structure $Y$.}
    \label{fig:llm_back_parsing}
\end{figure}

However, this approach indirectly uses LLM for treebank annotation, not taking full advantage of LLM abilities in domain generalization and language comprehension.
Besides, the two-stage pipeline inevitably introduces noise and error propagation, resulting in parse trees with errors and limited cross-domain constituency transfer.
Compared with an unlabeled raw corpus, guiding LLMs to directly generate the labeled constituency treebank in the target domain could be a more direct and effective approach for cross-domain constituency parsing. 
Concretely, direct treebank generation resorts to powerful LLMs to generate the target domain sentence and the corresponding constituency tree simultaneously, which can directly train a cross-domain constituency parser.

To this end, we explore how to utilize LLMs to generate a cross-domain constituency treebank effectively in this paper.
Since automatic treebank annotation by LLMs on the unlabeled sentence has poor performance~\citep{bai2023constituency}, we propose a novel treebank generation method as shown in \figurename~\ref{fig:llm_back_parsing}, LLM back generation, which is similar to the reverse process of constituency parsing.
\citet{yang2022challenges} indicate that both syntactical structure and domain vocabulary are crucial influence factors for cross-domain constituency parsing.
Therefore, LLM back generation builds the cross-domain constituency treebank by filling in missing sentential words based on the target domain constituency tree structure and domain keywords in the tree.
Concretely, we first extract the constituency tree and domain keywords on the target domain sentence.
Then we reserve domain keywords and remove other sentential words from the target domain constituency tree, which implies the character of the target domain on the syntactical structure and domain vocabulary.
Finally, we supply the LLM with the masked constituency tree and guide it to output the complete cross-domain parse tree.

In order to alleviate the noise in the LLM back generation treebank and reduce the cost and scale of LLM treebank generation, we design a span-level contrastive learning pre-training strategy for cross-domain constituency parsing, which can expand pre-training data significantly.
For each constituent span, the span-level contrastive learning pre-training distinguishes the related valid constituent spans and invalid spans with adjacent boundaries.
Specifically, we mine the left child, right child, parent and brother nodes as positive instances and the corresponding fifteen invalid spans as negative instances.
To the best of our knowledge, we are the first to introduce contrastive learning into constituency parsing.

We conduct experiments to verify the effectiveness of our LLM back generation treebank and contrastive learning pre-training strategy for cross-domain constituency parsing.
The news-domain constituency treebank PTB~\cite{marcus-etal-1993-building} is selected as the source, and a multi-domain constituency treebank MCTB~\cite{yang2022challenges} as the target, which consists of five domains.
Experimental results show that the LLM back generation treebank coupled with contrastive learning pre-training achieves state-of-the-art cross-domain parsing performance on the average F1 score, outperforming various baselines, including natural corpus treebank, conventional parsers, masked language modeling pre-training, previous cross-domain methods and large language models\footnote{Our code is public on \url{https://github.com/guopeiming/Back_Parsing_LLM}}.

\section{Related Work}

\paragraph{Cross-domain Constituency Parsing.}
Constituency parsing is an important and fundamental task in computational linguistics, which has not been completely solved.
The main challenge is stable cross-domain parsing performance.
Early work of constituency parsing focuses on the news domain \citep{collins-1997-three,stern2017minimal} and short sentences \citep{mcclosky-etal-2006-effective,mcclosky-etal-2008-self}.
In recent years, the natural language processing community has begun to pay attention to constituency parsing on different domains.
So there has been limited work investigating cross-domain constituency parsing.
\citet{mcclosky2010automatic} propose multiple source parser adaptation, which trains constituency parsers on multiple domain treebanks and combines these models by linear regression.
\citet{joshi2018extending} study single source domain adaptation based on the contextualized word representations, where they train the parsers on PTB only for similar target domains.
For syntactically distant target domains, they employ a dozen partial annotations to improve cross-domain constituency parsing performance.
\citet{fried-etal-2019-cross} and \citet{yang2022challenges} perform a systematic analysis on various constituency parsers.
\citet{yang2022challenges} annotate a constituency treebank MCTB, which contains five target domains.
\citet{guo2024cross} improve cross-domain constituency parsing performance by leveraging heterogeneous data from different types of tasks.

Researchers have investigated the effect of LLMs on cross-domain constituency parsing in recent years.
\citet{bai2023constituency} conduct a comprehensive experiment on various LLMs, including ChatGPT, GPT-4, OPT, LLaMA and Alpaca.
They also explore the influence of different linearizations and LLM settings, including zero-shot, few-shot and fine-tuning.
\citet{li-etal-2023-llm} use grammar rules and target domain sentences as the input restriction and reference to guide ChatGPT to generate raw corpora.
Our work is this line since we also focus on LLMs and cross-domain constituency parsing.
However, we propose LLM back generation, which exploits the incomplete cross-domain constituency tree as the input restriction and generates the full parse tree for cross-domain constituency parsing.

\begin{figure*}[ht]
\centering
\includegraphics[width=\textwidth]{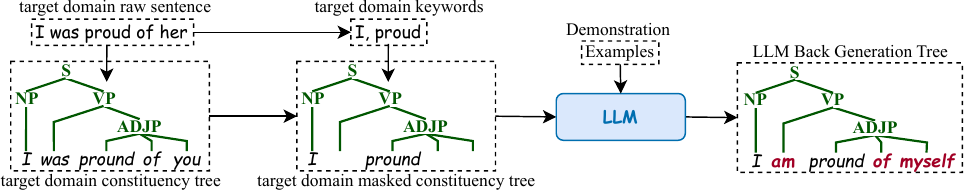}
\caption{Overview of LLM Back Generation. We first extract the target domain constituency tree and domain keywords. Then we mask all sentential words except domain keywords from the constituency tree. Finally, the LLM back generates the whole syntax tree based on the masked tree and some demonstration examples.}
\label{fig:llm_back_gen}
\end{figure*}

\paragraph{Contrastive Learning.}
Contrastive learning~\citep{chopra2005learning,schroff2015facenet,sohn2016improved} is first proposed and widely applied in the computer vision community~\citep{He_2020_CVPR,pmlr-v119-chen20j,caron2020unsupervised}.
Then researchers introduce contrastive learning into various natural language processing tasks, including sentence representation~\citep{gao-etal-2021-simcse}, event extraction~\citep{wang-etal-2021-cleve}, retrivel~\citep{yang-etal-2023-longtriever,zhang2023language}, vision-language pre-training~\citep{pmlr-v139-radford21a,singh-etal-2023-coarse}, etc.
To our best knowledge, we are the first to introduce contrastive learning into constituency parsing.
Concretely, contrastive learning is a technique that pulls similar examples (positive instances) and pushes different examples (negative instances).
In this work, we adopt contrastive learning on the span level to distinguish valid constituent spans and invalid spans, which can expand pre-training data and reduce the cost of LLM back generation significantly.
Besides, contrastive learning usually builds positive instances by data augmentation (e.g., image rotation or crop, token dropout or sentence paraphrase), and negative instances by other examples in the same batch.
However, our proposed strategy takes the left child, right child, parent and brother nodes as positive instances, and the fifteen corresponding invalid spans as negative instances.

\section{Method}

In this section, the process of LLM back generation (\S~\ref{text:llm_back_parsing}) is first introduced to generate the cross-domain constituency treebank.
Then we describe the chart-based parser~\citep{kitaev-klein-2018-constituency} briefly (\S~\ref{text:constituency_parser}).
Finally, based on the parser, we propose the contrastive learning pre-training strategy (\S~\ref{text:contrastive_learning_pretraining}), which acquires a better constituent span representation model by the LLM back generation treebank for cross-domain constituency parsing.

\subsection{LLM Back Generation}
\label{text:llm_back_parsing}

Constituency treebank $\{(X, Y)\}$ is crucial for a high-performance constituency parser, where $X$ and $Y$ are the input sentence with $n$ words $X = x_1 \cdots x_n$ and the corresponding constituency parse tree, respectively.
Treebank annotation can be extremely expensive and time-consuming, and only a few domains have large annotated treebanks.
Although LLMs can substitute human labor to annotate datasets in some natural language processing tasks~\citep{zhang-etal-2023-llmaaa,bansal2023large}, their performance on elaborate structure tasks like constituency parsing is poor~\citep{bai2023constituency}, underperforming conventional chart-based parsers.
One possible reason may be that LLMs are trained for dialogue rather than structure extraction.
Consequently, hallucinations in autoregressive generation make it difficult for the generated syntax trees to conform to the constraints of high-quality, valid tree structures.
Additionally, LLMs can not find strict annotation specifications from limited demonstration samples.
Therefore, we propose LLM back generation to alleviate these challenges in LLM treebank annotation.
As illustrated in~\figurename~\ref{fig:llm_back_gen}, LLM back generation takes the incomplete cross-domain constituency tree as input and fills masked words, which can ensure syntactic structure validity from the input end.

\paragraph{Cross-domain constituency tree preparation.}
As the reference, constraint and guidance, the incomplete target domain constituency tree is crucial for the LLM to generate the effective full parse tree.
\citet{yang2022challenges} indicate that both syntactic structure and domain vocabulary are important influence factors for cross-domain constituency parsing.
Therefore, we prepare the target domain masked constituency parse tree from these two aspects by extracting the bare constituency tree of the target domain raw sentence (syntactic structure) and removing all sentential words except target domain keywords (domain vocabulary).
\figurename~\ref{fig:llm_back_gen} shows the target domain masked constituency parse tree.

For cross-domain syntactic structure, we first train the state-of-the-art chart-based parser~\citep{kitaev-klein-2018-constituency} and then parse the constituency tree corresponding to the unlabeled target domain raw sentence.
Generally, the output constituency syntax tree will imply some cross-domain syntactical tree structures, since the raw sentence is derived from the target domain.

For domain vocabulary, we first extract domain keywords from the sentence based on the KeyBERT~\citep{grootendorst2020keybert}, which computes all embedding similarities between words and the sentence and selects topK words as keywords.
Then we retain 25\% sentential words that are the most similar to the origin sentence as domain keywords and remove the left words from the constituency parse tree.
In particular, our approach is equivalent to \citet{li-etal-2023-llm} when all words are reserved, while domain vocabulary can not be controlled when all words are masked.

\paragraph{LLM back generation.}
As shown in \figurename~\ref{fig:llm_back_gen}, LLM back generation generates the full constituency syntax tree by in-context-learning (ICL)~\citep{brown2020language}.
First, the target domain incomplete constituency syntax tree is fed to the LLM.
Second, we also input some demonstration examples, which help the LLM to better comprehend the treebank back generation task.
Concretely, the demonstration example is a pair of masked and full constituency trees, which are derived from the target domain raw corpus.
We append a detailed prompt string of LLM back generation in \S~\ref{text:app:llm_back_gen_prompt}.
Third, the LLM fills the masked syntax tree with appropriate words to generate a new sentence $\widehat{X}$ conforming the constituency parse tree structure $Y$, imitating demonstration $D$:
\begin{equation*}
    (\widehat{X}, Y) = \text{LLM}(Y, X^{*}, D),
\end{equation*}
where $X^{*}$ denotes the masked sentence, and the tuple $(\widehat{X}, Y)$ is the LLM back generation parse tree, which implies domain characteristics.

\subsection{Chart-based Constituency Parser}
\label{text:constituency_parser}
We briefly introduce the chart-based constituency parser~\citep{kitaev-klein-2018-constituency}, which is the foundation of our span-level contrastive learning pre-training.
Concretely, the parser first adopts BERT~\citep{devlin2018bert} to vectorize the input sentence, then encodes the context information by a partitioned transformer, and computes span representations $\bm{r}$ with a span encoder~\footnote{Parser details refer to \citet{kitaev-klein-2018-constituency}.}.
Subsequently, a multi-layer perceptron assigns a score $s(i, j, l)$ to each labeled span, which represents the score of the span as a constituent with the syntactic label $l$.
Finally, the score of the constituency tree $s(T)$ is computed by summing the scores of all the labeled spans within it.
Particularly, the chart-based parser exploits the CKY algorithm to efficiently search for the syntax tree with the highest score as the predicted output $\widehat{T}$.
For training, the tree-based max-margin loss minimizes the difference between the gold-standard tree $T^{*}$ and the predicted tree $\widehat{T}$:
\begin{equation*}
    \begin{aligned}
        \mathcal{L} = s(\widehat{T}) - s(T^{*}) + \varDelta (\widehat{T}, T^{*}),
    \end{aligned}
\end{equation*}
where $\varDelta$ represents the Hamming difference.

\subsection{Contrastive Learning Pre-training}
\label{text:contrastive_learning_pretraining}
In this subsection, we present our span-level contrastive learning pre-training strategy, which can make full use of the LLM back generation treebank.
For one thing, as pre-training is robust for the noise in the labeled datasets, we attempt to utilize the LLM back generation treebank to pre-train a remarkable constituent span representation model.
For another thing, the size of the LLM back generation treebank is limited due to the cost of LLM generation, thus our contrastive learning strategy is based on span-level not example-level, which can significantly expand pre-training data.
Specifically, we start with the introduction of the positive and negative instances for each constituent span $(i, j)$ in the LLM back generation tree.
The goal of the chart-based parser is to distinguish all valid constituent spans, thus the positive and negative instances are valid and invalid spans naturally.
Then, the contrastive constituent representation model is presented based on them.

\paragraph{Positive instances.}

\begin{figure}[t]
    \centering
\begin{subfigure}[b]{0.23\textwidth}
    \centering
    \includegraphics[width=\textwidth]{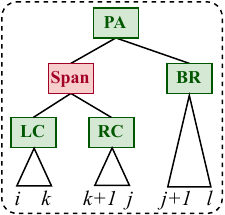}
    \caption{The constituent span as left sub-tree.}
    \label{fig:p_inst_l_tree}
\end{subfigure}
\hfill
\begin{subfigure}[b]{0.23\textwidth}
    \centering
    \includegraphics[width=\textwidth]{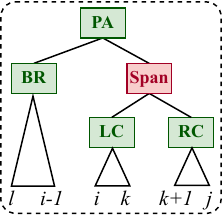}
    \caption{The constituent span as right sub-tree.}
    \label{fig:p_inst_r_tree}
\end{subfigure}
\caption{Positive instances (green node) and indexes for the constituent span (red node). LC, RC, PA and BR are left child, right child, parent, and brother, respectively.}
\label{fig:p_inst_tree}
\end{figure}

Non-local high-order features of upper and lower constituent nodes are essential for constituent recognition in the process of building a hierarchical parse tree~\citep{cui2021investigating,shi2022fast}.
As illustrated in \figurename~\ref{fig:p_inst_tree}, we select the left child (LC), right child (RC), parent (PA) and brother (BR) nodes as the positive instances for each constituent span.
Particularly, a constituency tree is a binary tree after binarization in the chart-based parser, so the constituent span is either a left sub-tree (\figurename~\ref{fig:p_inst_l_tree}) or a right sub-tree (\figurename~\ref{fig:p_inst_r_tree}).
When the constituent span $(i, j)$ is a left sub-tree, the set of the positive instances $(i, j)^{+}$ is $\{(i, k), (k+1, j), (i, l), (j+1, l)\}$.
When the constituent span $(i, j)$ is a right sub-tree, the set of the positive instances $(i, j)^{+}$ is $\{(i, k), (k+1, j), (l, j), (l, i-1)\}$.

\paragraph{Negative instances.}
In fact, span recognition in the chart-based parser is essentially boundary recognition.
Invalid spans with similar boundaries can impact the accuracy of recognition on constituent span $(i, j)$, such as $(i-1, j-1)$ and $(i+1, j+1)$.
Therefore, for each positive instance and constituent span itself, we mine corresponding negative instances, which are listed in \tablename~\ref{tab:n_inst}.
Specifically, the set of negative instances $(i, j)^{-}$ encompasses fifteen spans for constituent span $(i, j)$.
When the constituent span is a left or right sub-tree, twelve negative instances are the same but three are different.
Specially, some negative instances may be valid constituent spans in the constituency parse tree.
We filter these negative instances because we only need to pull all valid spans together and push all invalid spans apart.
Otherwise, it will destroy the contrastive constituent representation model.

\paragraph{Contrastive constituent representation model.}
Our contrastive constituent representation model is the same as the chart-based parser (\S~\ref{text:constituency_parser}), but replaces the multi-label classification layer and max-margin tree loss with contrastive objective only.
Therefore, after the contrastive constituent representation model is pre-trained on the LLM back generation treebank, we can transfer it into the constituency parser easily. 
For each span with the start index $i$ and end index $j$, our contrastive objective $\mathcal{L}$ is based on its span representation $\bm{r}$:
\begin{equation*}
\begin{aligned}
    \mathcal{L} = -\sum\limits_{m\in (i, j)^{+}} \text{log}\frac{e^{f(\bm{r}, \bm{r}^{+}_{m})}}{\sum\limits_{n \in (i, j)^{-}}e^{f(\bm{r}, \bm{r}^{+}_{m})} + e^{f(\bm{r}, \bm{r}^{-}_{n})}},
\end{aligned}
\end{equation*}
where $f$ indicates the cosine similarity function divided by temperature factor $\tau$.

\begin{table}[t]
\centering

\begin{tabular}{c|c|c}

\toprule
\multirow{2}{*}{\textbf{PosInst}} & \multicolumn{2}{c}{\textbf{NegInst}} \\
\cline{2-3}
& \textbf{left sub-tree} & \textbf{right sub-tree}\\
\hline
Span & \multicolumn{2}{c}{($i$, $j$$\pm$1), ($i$$\pm$1, $j$), ($i$$\pm$1, $j$$\pm$1)} \\
LC & \multicolumn{2}{c}{($i$, $k$-1), ($i$, $k$+1)} \\
RC & \multicolumn{2}{c}{($k$, $j$), ($k$+2, $j$)} \\
\hline
PA & ($i$, $l$-1), ($i$, $l$+1) & ($l$-1, $j$), ($l$+1, $j$) \\
BR & ($j$, $l$) & ($l$, $i$) \\

\bottomrule
\end{tabular}

\caption{Negative instances for the constituent span.}
\label{tab:n_inst}
\end{table}

We pre-train BERT~\citep{devlin2018bert}, partitioned transformer and span encoder by span-level contrastive learning on the LLM back generation treebank, and get the contrastive constituent representation model, which can generate a better span representation for cross-domain constituency parsing.
After pre-training, we equip the constituent representation model with max-margin tree loss and fine-tune it on the combination of the limited source domain human annotation treebank and the target domain LLM back generation treebank.

\section{Experiments}

\begin{table*}[t]
\centering
\setlength{\tabcolsep}{9pt}

\begin{tabular}{c|c |ccc cc c}

\toprule
\textbf{Method} & \textbf{Option} & \textbf{Dia} & \textbf{For} & \textbf{Law} & \textbf{Lit} & \textbf{Rev} & \textbf{Avg.}\\

\hline
\multicolumn{8}{c}{\emph{Large Language Models}} \\
\hline

\multirow{2}{*}{ChatGPT} & full & 30.54 & 18.86 & 24.93 & 11.96 & 28.67 & 22.99 \\
 & valid & 70.38 & 70.36 & 80.70 & 74.74 & 69.08 & 73.05 \\
\hline
\multirow{2}{*}{GPT-4} & full & 37.89 & 25.73 & 31.25 & 18.06 & 33.71 & 29.33 \\
 & valid & 77.64 & 76.27 & 84.49 & 79.58 & 75.63 & 78.72 \\

\hline
\multicolumn{8}{c}{\emph{Ours}} \\
\hline

\multirow{3}{*}{\shortstack{Natural Corpus \\ Treebank}} 
& DAPT & 86.25 & 87.04 & 92.19 & 86.42 & 83.86 & 87.15 \\
& NOPT & 86.54 & 87.47 & 92.26 & 86.64 & 83.98 & 87.38 \\
& CTPT & 87.33 & 87.80 & 92.54 & 86.91 & 84.35 & 87.79  \\ 

\hline

\multirow{3}{*}{\shortstack{LLM Back Generation \\ Treebank}} 
& DAPT & 86.50 & 87.28 & 92.43 & 86.71 & 84.02 & 87.39 \\
& NOPT & 87.75 & 87.43 & 92.57 & 87.01 & 84.28 & 87.81 \\
& CTPT & \textbf{87.92} & \textbf{88.13} & 93.22 & 87.50 & \textbf{85.86} & \textbf{88.52} \\

\hline
\multicolumn{8}{c}{\emph{Previous Work}} \\
\hline
\citet{kitaev-klein-2018-constituency} & -- & 86.10 & 86.92 & 92.07 & 86.28 & 84.32 & 87.14 \\
\citet{liu2017order} & -- & 85.56 & 86.33 & 91.50 & 84.96 & 83.89 & 86.45 \\
\citet{li-etal-2023-llm} & -- & 87.59 & 87.55 & \textbf{93.29} & \textbf{87.54} & 85.58 & 88.31\\

\bottomrule
\end{tabular}

\caption{
Main results on MCTB benchmark.
DAPT, NOPT and CTPT are short for domain adaptive pre-training, no pre-training, and our contrastive learning pre-training, respectively.
}
\label{tab:main_res}
\end{table*}

\subsection{Experimental Setup}

\paragraph{Datasets and Hyperparameters.}
Following~\citet{li-etal-2023-llm}, we use MCTB~\citep{yang2022challenges} as the target constituency parsing dataset, which includes five target domains: dialogue (Dia), forum (For), law, literature (Lit) and review (Rev).
Based on \texttt{gpt-4-1106-preview}~\citep{openai2023gpt}, we generate 10,000 constituency trees as the LLM back generation treebank containing the above five domains.
We attempt other large language models (e.g., ChatGPT and Llama-3) as well, but they mostly either fail to generate constituency trees or produce trees with errors.
\tablename~\ref{tab:llm_res} in \ref{text:app:llm_res} reports their results.
Besides, the other dataset details and the hyperparameters of proposed contrastive learning are also placed in~\ref{text:app:hyper} for limited space.

\paragraph{Evaluation.}
F1 score of labeled bracketed spans is used to evaluate the performance of cross-domain constituency parsing.
We conduct the experiments on three different random seeds and report the average results, ignoring punctuations following~\citet{kitaev-klein-2018-constituency} and \citet{li-etal-2023-llm}.

\paragraph{Baselines.}
We compare our approach with various constituency parsers:
(1) a strong Transition-based constituency parser~\citep{liu2017order}, (2) a strong chart-based constituency parser~\citep{kitaev-klein-2018-constituency}, which is re-implemented by us as the basic constituency parser and (3) a state-of-the-art cross-domain constituency parsing method~\citep{li-etal-2023-llm}, which utilizes the LLM to generate unlabeled raw sentences in the target domain.

We also report the cross-domain constituency parsing performances of \emph{ChatGPT}~\citep{brown2020language,ouyang2022training} and \emph{GPT-4}.
We use \texttt{gpt-3.5-turbo} and \texttt{gpt-4} to generate bracketed parse trees with in-context-learning \citep{brown2020language}, where 10 constituency tree examples from the source treebank PTB are prepended before the testing instance as demonstrations.
Notably, the outputs can contain numerous errors, including unmatched brackets, omitted words from input sentences, and responses lacking bracketed parse trees, because LLMs predict the next token auto-regressively and do not guarantee the validity of generated constituency trees.
We report results both considering and not considering invalid trees in the lines of ``full'' and ``valid'' in \tablename~\ref{tab:main_res}.

For the contrastive learning pre-training (CTPT) strategy, we compare it with two pre-training methods: (1) no pre-training (NOPT) directly fine-tunes the chart-based parser on the source treebank and LLM back generation treebank. (2) domain adaptive pre-training (DAPT)~\citep{gururangan-etal-2020-dont} continues pre-training BERT from \texttt{BERT-large-uncased} on the target domain raw corpus, which comprises 500k sentences in the five target domains, 100k sentences for each domain.

In order to verify the effectiveness of our LLM back generation treebank, we conduct a contrast experiment on the \emph{Natural Corpus Treebank}, which is parsed from the natural target domain raw corpora by a basic parser~\citep{kitaev-klein-2018-constituency}.

\subsection{Main Results}

\tablename~\ref{tab:main_res} reports F1 scores of different cross-domain constituency parsing methods on MCTB, which consists of five targe domains.

First, we examine the parsing performances of large language models.
\emph{ChatGPT} and \emph{GPT-4} show poor performance on all domains, which suggests that such generative LLMs can be less capable of solving cross-domain constituency parsing.
Besides, we find that ChatGPT and GPT-4 tend to generate invalid parse trees.
Take the input sentence ``\emph{He is right .}'' for example, LLMs might generate unmatched brackets (e.g., \emph{``[S [NP [PRP He]] [VP [VBD was] [ADJP [JJ right] [. .]''}) or drop sentential words (e.g., \emph{``[S [VP [VBD was] [ADJP [JJ right]]] [. .]]''}).
Therefore, when taking all outputs (the second line ``full'') into evaluation, their performances decrease severely.
The poor parsing results and invalid parse tree prove that it is hard for LLMs to annotate treebank directly.

Second, we look at the performances of LLM back generation treebank and natural corpus treebank.
As shown in \tablename~\ref{tab:main_res}, our LLM back generation treebank can boost the parsing results significantly, leading the gains on the average F1 score by $87.39-87.15=0.24$, $87.81-87.38=0.67$ and $88.52-87.79=0.73$ for three pre-training methods, respectively.
Regardless of which pre-training method is used, the performance of our LLM back generation treebank is significantly superior to that of natural corpus treebank, which demonstrates the effectiveness of LLM back generation treebank.

Third, we observe the results of different pre-training strategies.
DAPT is inferior to NOPT and close to basic chart-based parser~\citep{kitaev-klein-2018-constituency}, which is trained only in the source treebank.
One reason could be that the pre-training format of masked language modeling is far from constituency parsing.
Although the model is trained on the target domain corpora, the knowledge is hard to transfer to constituent recognition.
CTPT significantly improves the parsing performance across five domains compared with DAPT and NOPT.
The observation suggests that our span-level contrastive learning pre-training effectively acquires constituent knowledge from the LLM back generation treebank.

Finally, we make comparisons with the previous work.
Our LLM back generation treebank coupled with span-level contrastive learning pre-training obtains state-of-the-art cross-domain parsing results on the average F1 score.
Compared with raw corpus generation from LLM~\citep{li-etal-2023-llm}, our approach gains the stronger average F1 score, which shows the effectiveness of LLM back generation treebank generation.
For the dialogue, forum and review domain, we perform a statistical significance analysis between our approach and~\citet{li-etal-2023-llm}, where $p < 0.05$ verifies the effectiveness of our proposed method.
For the law and literature domain, we suspect the reason for lower results might be differences in domain distribution and sentence length.
The law and literature domains usually involve more formal and official expressions and long sentences, while more colloquial and short sentences exist in the dialogue, forum and review domains.
For one thing, improving the performance of long sentences may be harder than short sentences.
For another thing, our method handles all five domains by only one parser, which needs to balance different domain differences, however, \citet{li-etal-2023-llm} train a separate parser for each domain, which only considers sentences from the same domain with a similar distribution.
More analyses and experiments are in \ref{text:app:res_comp}.

\subsection{Analyses}
We conduct detailed experimental analyses in this subsection to offer several findings of our LLM back generation treebank and span-level contrastive learning pre-training.

\begin{figure}
\centering

\begin{tikzpicture}

\begin{axis}[
    ybar,
    xmin=-1.1,
    xmax=11.1,
    ymin=87.00,
    ymax=88.50,
    width=8.3cm,
    height=4.2cm,
    bar width=8,
    ylabel={F1 score},
    y label style={font=\footnotesize, yshift=-0.75cm},
    ytick={87.1, 87.7, 88.3},
    yticklabels={87.1, 87.7, 88.3},
    yticklabel style={font=\scriptsize, xshift=0.05cm, rotate=90},
    xlabel={mask rate of sentential word},
    x label style={font=\footnotesize, yshift=0.25cm},
    xtick={0, 2.5, 5.0, 7.5, 10.0},
    xticklabels={0\%, 25\%, 50\%, 75\%, 100\%},
    xticklabel style={font=\scriptsize, yshift=0.15cm},
    legend style={at={(0.778, 1.0)}, anchor=north, legend columns=3, font=\tiny, inner sep=0.8pt, outer sep=0pt, column sep=-0.5pt},
]

    \addplot[r_d, fill=r_d, bar shift=0pt, pattern=north east lines, pattern color=r_d] coordinates {
        (-0.6, 87.15)
    };
    \addplot[g_d, fill=g_d, bar shift=0pt, pattern=grid, pattern color=g_d] coordinates {
        (0, 87.38)
    };
    \addplot[b_d, fill=b_d, bar shift=0pt, pattern=north west lines, pattern color=b_d] coordinates {
        (0.6, 87.79)
    };

    \addplot[r_d, fill=r_d, bar shift=0pt, pattern=north east lines, pattern color=r_d] coordinates {
        (1.9, 87.39)
    };
    \addplot[g_d, fill=g_d, bar shift=0pt, pattern=grid, pattern color=g_d] coordinates {
        (2.5, 87.82)
    };
    \addplot[b_d, fill=b_d, bar shift=0pt, pattern=north west lines, pattern color=b_d] coordinates {
        (3.1, 88.48)
    };

    \addplot[r_d, fill=r_d, bar shift=0pt, pattern=north east lines, pattern color=r_d] coordinates {
        (4.4, 87.30)
    };
    \addplot[g_d, fill=g_d, bar shift=0pt, pattern=grid, pattern color=g_d] coordinates {
        (5.0, 87.67)
    };
    \addplot[b_d, fill=b_d, bar shift=0pt, pattern=north west lines, pattern color=b_d] coordinates {
        (5.6, 88.205)
    };

    \addplot[r_d, fill=r_d, bar shift=0pt, pattern=north east lines, pattern color=r_d] coordinates {
        (6.9, 87.25)
    };
    \addplot[g_d, fill=g_d, bar shift=0pt, pattern=grid, pattern color=g_d] coordinates {
        (7.5, 87.53)
    };
    \addplot[b_d, fill=b_d, bar shift=0pt, pattern=north west lines, pattern color=b_d] coordinates {
        (8.1, 88.10)
    };

    \addplot[r_d, fill=r_d, bar shift=0pt, pattern=north east lines, pattern color=r_d] coordinates {
        (9.4, 87.20)
    };
    \addplot[g_d, fill=g_d, bar shift=0pt, pattern=grid, pattern color=g_d] coordinates {
        (10.0, 87.43)
    };
    \addplot[b_d, fill=b_d, bar shift=0pt, pattern=north west lines, pattern color=b_d] coordinates {
        (10.6, 87.86)
    };

    \legend{DAPT, NOPT, CTPT};

\end{axis}

\end{tikzpicture}

\caption{F1 score of three pre-training strategies on the LLM back generation treebanks with different mask rates of sentential word.}
\label{fig:word_mask}
\end{figure}
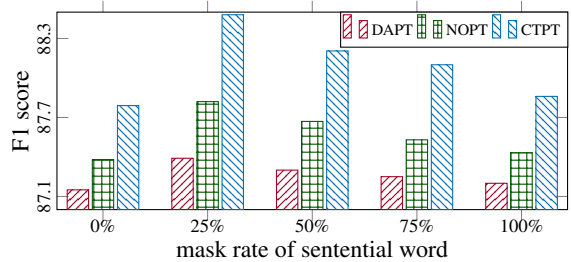

\paragraph{Mask rate for LLM back generation treebank.}
The mask rate of sentential words in the target constituency tree may influence LLM back generation treebank generation, which is the core of our approach.
We conduct an experiment to examine the relation between cross-domain constituency parsing performance and mask rate for LLM back generation treebank.
\figurename~\ref{fig:word_mask} shows the results based on the average F1 score of five target domains.
When the mask rate is 0\%, all sentential words are reserved, thus it is the natural corpus treebank in fact.
When the mask rate is 100\%, all sentential words are masked, which exploits LLM to generate a treebank based on the bare constituency tree only.

First, we can see that all LLM back generation treebanks including 25\%, 50\% and 100\% mask rates outperform the natural corpus treebank.
This phenomenon shows the effectiveness of our LLM back generation treebank.
Second, the 25\% mask rate achieves the best F1 score.
As the mask rate becomes larger, the cross-domain parsing performance decreases gradually.
A reasonable explanation might be that fewer retained domain keywords cause LLM generation to be freer.
Therefore, the final LLM back generation treebank will shift from the target domain.
Third, our contrastive learning pre-training strategy significantly improves the results compared with DAPT and NOPT in all settings, which verifies its effectiveness.

\begin{figure}
\centering

\begin{tikzpicture}

\begin{axis}[
    ymax=88.5,
    ymin=0,
    xmax=10,
    xmin=0,
    ylabel={F1 score},
    y label style={font=\footnotesize, yshift=-0.8cm},
    ytick={0, 20, 40, 60, 80},
    yticklabels={0, 20, 40, 60, 80},
    y tick label style = {font=\scriptsize, xshift=0.1cm,},
    xlabel={Training Step},
    x label style={font=\footnotesize, yshift=0.3cm},
    xtick = {0,1,2,3,4,5,6,7,8,9, 10},
    xticklabels={0, 100, 200, 300, 400, 500, 600, 700, 800, 900, 1000},
    x tick label style = {font=\scriptsize, yshift=0.1cm},
    width=8.3cm,
    height=4.2cm,
    legend style={at={(0.835, 0.6)}, anchor=north, legend columns=1, font=\scriptsize},
]

    \addplot[thick, solid, r_d, mark=square*, mark size = 1.2pt, mark options={fill=r_d, solid}] coordinates{
        (0, 0) (1, 40) (2, 60) (3, 69) (4, 75) (5, 80) (6, 82) (7, 83) (8, 83.8) (9, 84.4) (10, 84.4)
    };\addlegendentry{CTPT}

    \addplot[thick, solid, g_d, mark=|, mark size = 2.5pt, mark options={fill=g_d, solid}] coordinates{
        (0, 0) (1, 14) (2, 30) (3, 42) (4, 53) (5, 62) (6, 69) (7, 75) (8, 79) (9, 81) (10, 82)
    };\addlegendentry{NOPT}

    \addplot[thick, solid, b_d, mark=o, mark size = 1.8pt, mark options={fill=b_d, solid}] coordinates{
        (0, 0) (1, 12) (2, 25) (3, 38) (4, 49) (5, 59) (6, 66) (7, 71) (8, 75) (9, 79) (10, 80)
    };\addlegendentry{DAPT}

\end{axis}
  
\end{tikzpicture}

\caption{Convergence curve of different pre-training strategies on the LLM back generation treebank.}
\label{fig:convergence}
\end{figure}
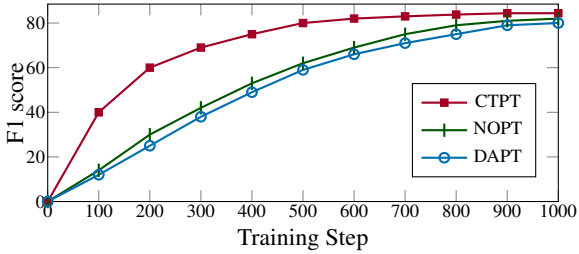

\paragraph{Convergence for contrastive learning pre-training.}
Contrastive learning pre-training is also one of the key contents of this paper.
We find that convergences of different pre-training strategies are different in the fine-tuning phase.
Specifically, we evaluate the cross-domain constituency parsing performance every 100 training steps when pre-trained models are fine-tuned on the constituency treebank.
The results are illustrated in \figurename~\ref{fig:convergence}, where the x-axis denotes the training step and the y-axis denotes the average F1 score on all five target domains.
When the training step is zero, the F1 scores of three pre-training methods are zero, because the parsers are only pre-trained not fine-tuned on the treebank.
As the training step grows larger in the initial phase, the three curves increase significantly.
Afterward, the performance stops increasing and gradually converges.

\figurename~\ref{fig:convergence} illustrates that our contrastive learning pre-training converges significantly faster compared with domain adaptive pre-training and no pre-training.
Concretely, CTPT converges at 600 training steps while DAPT and NOPT stop increasing after 1000 training steps.
Besides, the curve of CTPT not only improves fast but also gains the best results finally.
The observation suggests that our span-level contrastive learning pre-training strategy pre-trains the contrastive constituency representation model that can compute better span representations.
Our proposed pre-training acquires and transfers the knowledge of constituent recognition into cross-domain constituency parsing successfully.

\begin{figure}
\centering

\begin{tikzpicture}

\begin{axis}[
    ymax=88.52,
    ymin=87.33,
    xmax=10,
    xmin=0,
    ylabel={F1 score},
    y label style={font=\footnotesize, yshift=-0.7cm},
    ytick={87.5, 88.1},
    yticklabels={87.5, 88.1},
    y tick label style = {font=\scriptsize, xshift=0.0cm, rotate=90},
    xlabel={Treebank Size},
    x label style={font=\footnotesize, yshift=0.3cm},
    xtick = {0,2,4,6,8,10},
    xticklabels={0, 2k, 4k, 6k, 8k, 10k},
    x tick label style = {font=\scriptsize, yshift=0.1cm},
    width=8.0cm,
    height=4.2cm,
    legend style={at={(0.687, 0.72)}, anchor=north, legend columns=1, font=\scriptsize, inner sep=0.0pt, outer sep=0pt, row sep=-1.5pt},
]

    \addplot[thick, solid, r_d, mark=square*, mark size = 1.2pt, mark options={fill=r_d, solid}] coordinates{
        (0, 87.81) (1, 87.99) (2, 88.11) (3, 88.23) (4, 88.28) (5, 88.37) (6, 88.44) (7, 88.43) (8, 88.49) (9, 88.47) (10, 88.48)
    };\addlegendentry{LLM Back Generation Treebank}

    \addplot[thick, solid, g_d, mark=|, mark size = 2.5pt, mark options={fill=g_d, solid}] coordinates{
        (0, 87.38) (1, 87.44) (2, 87.53) (3, 87.57) (4, 87.60) (5, 87.69) (6, 87.74) (7, 87.76) (8, 87.78) (9, 87.77) (10, 87.78)
    };\addlegendentry{Natural Corpus Treebank}

\end{axis}
  
\end{tikzpicture}
\caption{F1 score of contrastive learning pre-training on natural corpus treebank and LLM back generation treebank with respect to the size of treebank.}
\label{fig:treebank_size}
\end{figure}
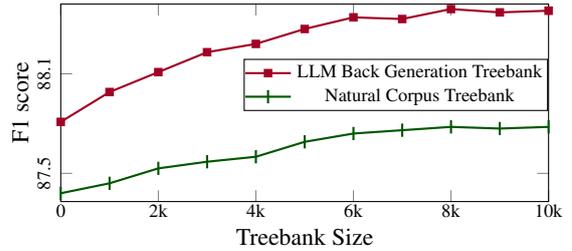

\paragraph{Contrastive learning pre-training treebank size.}
Intuitively, the final average cross-domain constituency parsing results (y-axis) should be dependent on the number of constituency trees extracted from natural corpus or LLM back generation (x-axis).
We conduct a experiment to observe the performance differences with respect to the size of contrastive learning pre-training treebank.
\figurename~\ref{fig:treebank_size} demonstrates the results.
When treebank size is 0, no constituency trees are used in contrastive learning pre-training, which is equivalent to NOPT.
The cross-domain parsing performance would stop increasing after 8k for both natural corpus treebank and LLM back generation treebank.
Although the size of pre-training treebank is limited, our proposed span-level contrastive learning pre-training can expand the limited sentence-level constituency trees to a large number of span-level spans.
Concretely, the average number of spans in the LLM back generation treebank is about 25.
Thus, our span-level contrastive learning pre-training is performed on $25*10,000 = 250,000$ examples, which significantly reduces the cost and scale of the LLM back parsing generation treebank.

\section{Conclusion}
In this paper, we proposed LLM back generation, which takes a masked cross-domain constituency tree as the input instruction and fills in missing words to generate the LLM back generation treebank.
Besides, we presented a span-level contrastive learning pre-training strategy for the LLM back generation treebank.
To the best of our knowledge, this is the first work to introduce contrastive learning into constituency parsing.
Experimental results showed that our LLM back generation treebank coupled with span-level contrastive learning pre-training gains state-of-the-art results on averaged F1 score compared with various baselines, including natural corpus treebank, conventional parsers, masked language modeling pre-training, previous cross-domain constituency parsing methods and large language models.


\section*{Limitations}
Our approach is verified on the only English dataset, so its effectiveness is not clear for the other languages.
To the best of our knowledge, there are no available cross-domain constituency treebanks for other languages, as dataset annotation is time-consuming and expensive, especially for parsing tasks.
So it is difficult to perform experiments on diverse languages to verify the effectiveness of the proposed method.
In fact, our approach can be applied directly to any language, and fortunately, GPT-4 is qualified in most languages.
In addition, when developers face a specific language, we also suggest trying other representative and popular LLMs for this language, such as Qwen or DeepSeek for Chinese, HyperCLOVA-X for Korean, and so on.

\section*{Acknowledgements}
We sincerely thank the reviewers for their invaluable feedback, especially Reviewer N7Sb in ARR 2024 December. This work was supported by the National Natural Science Foundation of China under Grant 61976180 and 62176180, and the Shenzhen Science and Technology Program under Grant GXWD20231130140414001 and ZDSYS20230626091203008.

\bibliography{custom}

\appendix

\section{Appendix}
\label{sec:appendix}

\subsection{LLM Back Generation}
\label{text:app:llm_back_gen_prompt}
\begin{figure*}[ht]
\centering

\begin{tcolorbox}[title=\textbf{LLM Back Generation Prompt},colbacktitle=gray!30,coltitle=black,colback=white, left=2mm, right=2mm]

\vspace{-6pt}
\textcolor{blue}{System Prompt:} 

You are a professional linguist.

\vspace{-5pt}\hdashrule[0.5ex]{15.2cm}{1pt}{4pt} \vspace{-5pt}

\textcolor{blue}{Demonstration:}

(SQ (VBP \textbf{\emph{Have}}) (NP (PRP )) (ADVP (DT )) (VP (VBN ) (NN \textbf{\emph{skiing}})))

(SQ (VBP \textbf{\emph{Have}}) (NP (PRP \textbf{\emph{you}})) (ADVP (DT \textbf{\emph{ever}})) (VP (VBN \textbf{\emph{gone}}) (NN \textbf{\emph{skiing}})))

\vspace{-5pt}\hdashrule[0.5ex]{15.2cm}{1pt}{4pt} \vspace{-5pt}

\textcolor{blue}{Masked Constituency Tree:}

(S (NP (PRP \textbf{\emph{I}})) (VP (VBD ) (ADJP (JJ \textbf{\emph{proud}}) (PP (IN ) (NP (PRP ))))))

\vspace{-7pt}
\end{tcolorbox}

\begin{tcolorbox}[title=\textbf{LLM Back Generation Output},colbacktitle=teal!30,coltitle=black,colback=white, left=2mm, right=2mm]

\vspace{-6pt}
\textcolor{blue}{LLM Back Generation Tree:}

(S (NP (PRP \textbf{\emph{I}})) (VP (VBD \textbf{\emph{am}}) (ADJP (JJ \textbf{\emph{proud}}) (PP (IN \textbf{\emph{of}}) (NP (PRP \textbf{\emph{myself}})))))

\vspace{-7pt}
\end{tcolorbox}

\caption{
The prompt and output of LLM back generation.
The blue texts are only shown for illustrative purposes and not in the actual prompt.}
\label{fig:llm_prompt}
\end{figure*}

\figurename~\ref{fig:llm_prompt} displays the prompt and output of LLM back generation.
Concretely, we first prompt the LLM as a professional linguist in the system prompt.
Then, the question and answer of the demonstration are appended to the system prompt.
The former is the incomplete constituency tree with both the syntactical structure and domain keywords of the target domain, and the latter is the original target domain full constituency syntax tree.
Following, the masked constituency tree to be processed is placed at the end of the input instruction.
Finally, the LLM fills the incomplete syntax tree with appropriate words to generate a new sentence conforming to the constituency parse tree structure, imitating the demonstration.

\subsection{Datasets and Hyperparameters}
\label{text:app:hyper}

We use PTB~\citep{marcus-etal-1993-building} and MCTB~\citep{yang2022challenges} as the source and target constituency parsing datasets, respectively.
Dataset statistics are listed in \tablename~\ref{tab:datasets}, where \#Sent and AS denote the number of sentences and average number of spans in binary constituency parse trees.
For the five target domain raw corpora, we collect unlabelled sentences with sources matching the corresponding target treebank in MCTB, including Wizard~\citep{dinan2018wizard}, Reddit~\citep{volske-etal-2017-tl}, ECtHR~\citep{DVN/OBYUO5_2019}, Gutenberg\footnote{\url{https://www.gutenberg.org/}}, and Amazon~\citep{he2016ups}.

For LLM back generation treebank generation, we set 2 demonstrations in the prompt for LLM back generation, which are placed in the \texttt{name} field of the system message.
For postags in unlabeled raw sentences, we train a BERT-LSTM-CRF model for prediction.
The training datasets are extracted from constituency treebank PTB~\citep{marcus-etal-1993-building} directly.
For keyword extraction, we use the default word representation model (i.e., \texttt{all-MiniLM-L6-v2}) of the KeyBERT python library \citep{grootendorst2020keybert}.
We randomly selected the sentential words to mask the sentence in preliminary experiments, but the results were limited as random masking would hurt the target vocabulary domain characteristics.
For contrastive learning pre-training, we use BERT-large-uncased as the pretrained language model backbone~\cite{devlin2018bert} for the cross-domain constituency parser and the other hyperparameters are the same as~\citet{kitaev-klein-2018-constituency}.
We use the AdamW algorithm with learning rate 3e-5, batch size 64 to pre-train LLM back generation treebank for 10 epochs.
For each constituency tree in a batch, only 20\% constituents are sampled as examples to compute the contrastive learning loss.
Temperature factor $\tau$ is 0.05. 
For constituency parsing fine-tuning, we use the AdamW algorithm with learning rate 1e-5, batch size 64, and linear learning rate warmup over the first 400 steps to optimize parameters.
We stop early training when the F1 score does not increase on the PTB development set for 4 epochs.
We attempt to mix the standard fine-tuning loss and the contrastive learning pre-training loss in preliminary experiments.
But there were no improvements and the training would unstable. So we use the standard loss only during the final fine-tuning stage. 
We run all experiments on a single Nvidia V100. 

\begin{table}[t]
    \centering
    \begin{tabular}{l l r r c c}
    
    \toprule
    \textbf{Dataset} & \textbf{Domain} & \textbf{\#Sent} & \textbf{AS} \\
    \midrule
    PTB & news & 39,832 & 25.54 \\

    \midrule
    \multirow{5}{*}{MCTB} 
    & dialogue & 1,000 & 15.92 \\
    & forum & 1,000 & 26.33 \\
    & law & 1,000 & 27.47\\
    & literature & 1,000 & 26.91 \\
    & review & 1,000 & 15.19 \\
    \bottomrule
    \end{tabular}

\caption{Datasets statistics.}
\label{tab:datasets}
\end{table}

\subsection{Results of Different LLMs}
\label{text:app:llm_res}
\begin{table*}[t]
\centering
\setlength{\tabcolsep}{9pt}

\begin{tabular}{c|c |ccc cc c}

\toprule
\textbf{LLM} & \textbf{Access} & \textbf{Dia} & \textbf{For} & \textbf{Law} & \textbf{Lit} & \textbf{Rev} & \textbf{Avg.}\\

\midrule

GPT-4 & close & \textbf{87.92} & \textbf{88.13} & \textbf{93.22} & \textbf{87.50} & \textbf{85.86} & \textbf{88.52} \\

ChatGPT & close & 87.63 & 87.91 & 93.03 & 87.29 & 85.56 & 88.28 \\

Llama-3 & open & 87.39 & 87.65 & 92.80 & 87.04 & 85.38 &  88.05 \\

\bottomrule
\end{tabular}

\caption{
Results on different LLMs.
}
\label{tab:llm_res}
\end{table*}

Generally, the quality of LLM back generation treebank is related to the used LLM.
We perform a experiment to analyse the results of different LLMs, which are reported in \tablename~\ref{tab:llm_res}.
We adapt extra two representative large language models, ChatGPT and Llama-3-8B~\citep{touvron2023llama}, to produce LLM back generation treebanks.
Then the proposed span-level contrastive learning strategy is conducted on the LLM back generation treebanks.
For the close-source model, ChatGPT generates more constituency trees with syntax errors.
For the open-source model, Llama-3 usually refuses to generate constituency trees, which may be short of pre-trained knowledge of constituency parsing.
Therefore, we fine-tune Llama-3 by LoRA~\citep{hu2022lora} on the target domain corpora with the task format in \figurename~\ref{fig:llm_prompt}.
Concretely, ChatGPT and Llama-3 achieve average F1 scores of 88.28 and 88.05 respectively, underperforming GPT-4.
The large language model with stronger understanding and generation capabilities can generate high-quality treebanks, leading to better cross-domain constituency parsing performance.

\subsection{Performance Comparison with Previous Method}
\label{text:app:res_comp}
\begin{table*}[t]
\centering

\begin{tabular}{c|ccc cc c}

\toprule
\textbf{Method} & \textbf{Dia} & \textbf{For} & \textbf{Law} & \textbf{Lit} & \textbf{Rev} & \textbf{Avg.}\\

\midrule

\citet{li-etal-2023-llm} & 87.59 & 87.55 & 93.29 & 87.54 & 85.58 & 88.31 \\
Ours &  \textbf{87.70} & \textbf{87.76} & \textbf{93.33} & \textbf{87.62} & \textbf{85.69} & \textbf{88.42} \\

\bottomrule
\end{tabular}

\caption{
Result comparison with \citet{li-etal-2023-llm} in the same settings.
}
\label{tab:res_comp}
\end{table*}

The method of \citet{li-etal-2023-llm} is closely related to ours, therefore we conduct detailed analyses and experiments for comparison here.
Compared with \citet{li-etal-2023-llm}, our method has other significant advantages except for a higher average parsing score in \tablename~\ref{tab:main_res}.
For one thing, our method needs only 10,000 sentences generated by LLMs, but \citet{li-etal-2023-llm} use ChatGPT to produce 200,000 sentences.
The size of the generated corpus is 20 times more than ours, which leads to an unaffordable and unavoidable high cost. We guess it is an important reason why they do not apply more powerful GPT-4 to generate sentences.
For another thing, our parser can handle sentences from all five domains simultaneously. But \citet{li-etal-2023-llm} train a separate parser for each domain.
In a real application scenario, our method deploys one model to parse different domains, while \citet{li-etal-2023-llm} need more models. This will waste valuable GPU resources.

Besides, a fair quantitative comparison between our method and \citet{li-etal-2023-llm} is important.
We run our model on the same settings as \citet{li-etal-2023-llm}: 1) using GPT-3.5-turbo, 2) scaling to 200,000 sentences, and 3) training a parser for each domain. \tablename~\ref{tab:res_comp} shows the experimental results.
Our method achieves better parsing performances on five domains, which verifies the effectiveness of our proposed LLM back generation treebank and span-level contrastive learning pretraining.
Besides, we also conduct statistical significance test experiments, where p < 0.05 shows the stability of experiments.

\subsection{Performance on Source Domain}
\label{text:app:res_sour}
\begin{table}[t]
\centering

\resizebox{\columnwidth}{!}{
\begin{tabular}{c|cccc cc c}

\toprule
\textbf{Model} & \textbf{Base chart parser} & \textbf{Our final method} \\

\midrule

F1 score & 95.64 & 95.71 \\

\bottomrule
\end{tabular}
}

\caption{
Results on the source domain.
}
\label{tab:res_sour}
\end{table}

Though our approach achieves impressive results on target domain, performances on source domain standard benchmarks are also important.
We test our final parser on the PTB dataset, which achieves an F1-score of 95.71 as shown in \tablename~\ref{tab:res_sour}. 
The result of the basic chart-based parser is 95.64, which is trained on the source domain PTB dataset.
The comparison verifies that our parser can maintain or even improve the performance on the standard benchmark.

\subsection{Example of Positive and Negative Instances}
\label{text:app:example_inst}
\begin{figure}[t]
    \centering
    \includegraphics[width=0.95\linewidth]{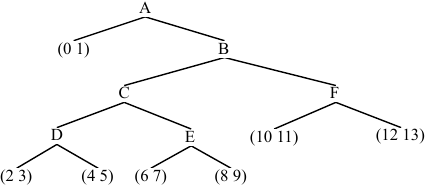}
    \caption{A constituency tree.}
    \label{fig:cont_tree}
\end{figure}
Given a constituency tree as illustrated in \figurename~\ref{fig:cont_tree}, uppercase letters (i.e., A, B, C, ...) are syntax labels, and numbers (i.e., 1, 2, 3, ...) are words.
Take the constituent node (C, 2, 9) for example, the positive instances are the left child (2, 5), right child (6, 9), parent (2, 13) and brother (10, 13).
The negative instances are invalid spans related to it: (1, 9), (3, 9), (2, 8), (2, 10), (1, 10), (3, 8), (1, 8), (3, 10), (2, 4), (2, 6), (5, 9), (7, 9), (2, 12), (2, 14), (9, 13).

\end{document}